\documentclass[10pt,twocolumn,letterpaper]{article}

\usepackage{cvpr}
\usepackage{times}
\usepackage{epsfig}
\usepackage{graphicx}
\usepackage{amsmath}
\usepackage{amssymb}
\usepackage{color, colortbl}

\usepackage{float}
\usepackage{subfig}

\usepackage[pagebackref=true,breaklinks=true,letterpaper=true,colorlinks,bookmarks=false]{hyperref}

\cvprfinalcopy 

\def\cvprPaperID{560} 

\definecolor{lightgray}{gray}{0.4}

\usepackage{xspace}
\usepackage{multirow}

\definecolor{rowCol}{rgb}{0.88,1,1}

\graphicspath{{imgs/}}

\ifcvprfinal\fi

\begin{document}

\title{Style Mixer: Semantic-aware Multi-Style Transfer Network.}

\author{Zixuan HUANG\thanks{indicates equal contribution.}\\
City University of Hong Kong\\
{\tt\small zixuan.huang@my.cityu.edu.hk }
\and
Jinghuai Zhang$^{*}$\\
City University of Hong Kong\\
{\tt\small jzhang538-c@my.cityu.edu.hk}
\and
Jing Liao\thanks{indicates corresponding author.}\\
City University of Hong Kong\\
{\tt\small jingliao@cityu.edu.hk}
}

\makeatletter
\let\@oldmaketitle\@maketitle
\renewcommand{\@maketitle}
{
\@oldmaketitle

 \centering
  \includegraphics[width=1.0\linewidth]
    {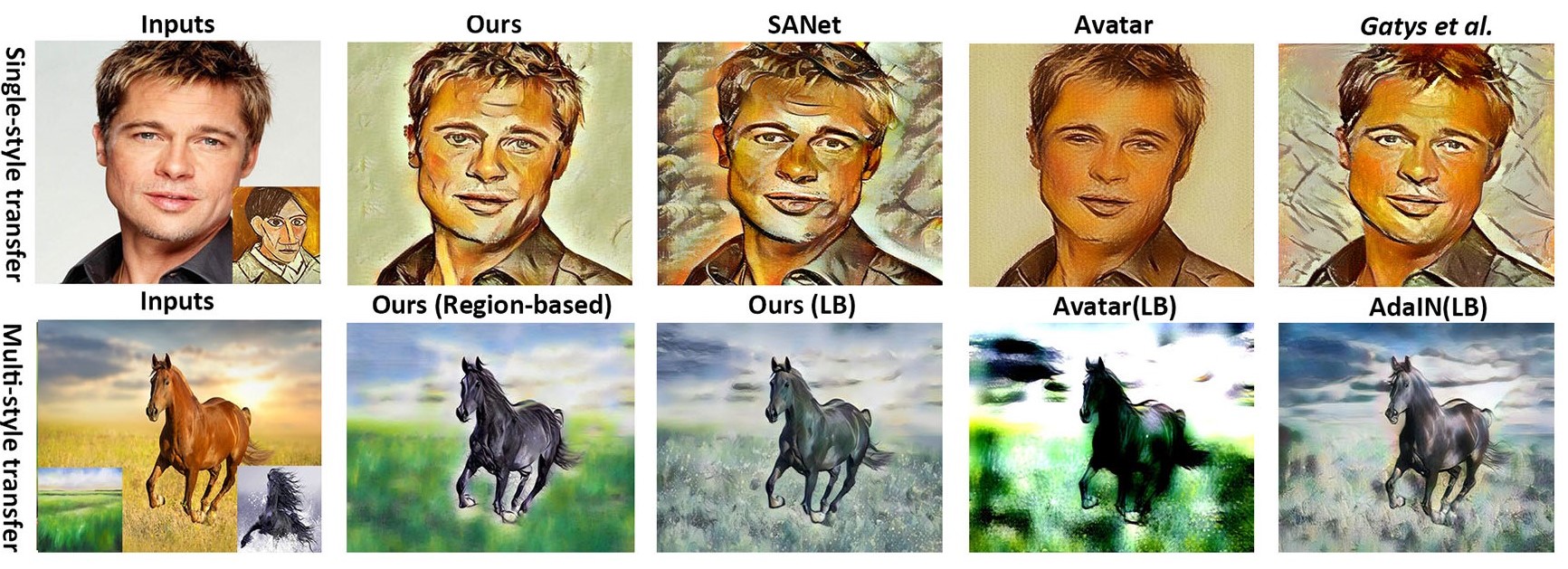}
  \captionof{figure}{In SST, our backbone model transfers style faithfully based on semantic correspondence. In MST, our proposed framework Style Mixer incorporates multiple styles based on regional semantics, therefore best preserve the characteristics of individual style and generates more favorable results than existing methods. LB indicates linear blending.}\label{fig:teaser1}
\bigskip
}

\makeatother

\maketitle
\thispagestyle{empty}

\begin{abstract}
Recent neural style transfer frameworks have obtained astonishing visual quality and flexibility in Single-style Transfer (SST), but little attention has been paid to Multi-style Transfer (MST) which refers to simultaneously transferring multiple styles to the same image. Compared to SST, MST has the potential to create more diverse and visually pleasing stylization results. In this paper, we propose the first MST framework to automatically incorporate multiple styles into one result based on regional semantics. We first improve the existing SST backbone network by introducing a novel multi-level feature fusion module and a patch attention module to achieve better semantic correspondences and preserve richer style details. For MST, we designed a conceptually simple yet effective region-based style fusion module to insert into the backbone. It assigns corresponding styles to content regions based on semantic matching, and then seamlessly combines multiple styles together. Comprehensive evaluations demonstrate that our framework outperforms existing works of SST and MST.
\end{abstract}

\section{Introduction}
The target of style transfer is to confer the style of a reference image to another image while preserving the content of the latter one. The seminal work of Gatys \etal \cite{gatys2015texture,gatys2016image} demonstrated that the correlation between the deep features is superior in capturing visual style. It opened up the era of neural style transfer. Later significant effort has been devoted to improving the speed, flexibility, and visual quality of neural style transfer. The most recent works \cite{huang2017arbitrary,li2017universal,li2018learning,chen2016fast,sheng2018avatar} support efficient arbitrary transfer style with a single convolutional neural network model, which serve as the state-of-the-art baselines.

However, most studies in neural style transfer focus on SST, i.e., the image is transferred by a single style reference. To generate more diverse and visually pleasing results, two straightforward attempts are proposed to extend existing techniques to MST, allowing the user to transfer the contents into an aggregation of multiple styles. One is linear blending \cite{gatys2016image,gu2018arbitrary,li2017universal,sheng2018avatar,park2018arbitrary,chen2016fast}, which interpolates features of different styles linearly by given weights. However, as is shown in Fig.~\ref{fig:teaser1}, this method tends to generate muddled results since the colors and textures of different styles are simply mixed, and also dull results since the combination is spatially invariant. Another method is to spatially combine multiple styles by asking users to provide a mask and manually assign the styles to different regions \cite{park2018arbitrary,li2017universal}, which results in the desired effect but involves tedious work.

In this paper, we propose a semantic-aware MST network: Style Mixer. It can automatically incorporate multiple styles into one result according to the regional semantics. Our Style Mixer consists of a backbone SST network and a multi-style fusion module. The backbone network can achieve semantic-level SST by learning the semantic correlations between the content and style features. It is inspired from two arbitrary style transfer networks: Avatar-Net \cite{sheng2018avatar} and SANet \cite{park2018arbitrary}. In order to build correspondences, Avatar-Net uses a fixed patch-swap module while SANet uses a learnable attention module. We incorporate the merits of both methods (leveraging patch information while allowing learnable parameters) by proposing a novel patch attention (PA) module for more accurate correspondences. PA improves traditional attention module by enabling the controllability of the size of the receptive field, which will benefit the works in other fields as well. Besides, we further improve the richness of style features by introducing multi-level feature fusion (MFF). Compared to the state-of-the-art style transfer networks, our backbone network is better in both capturing semantic correspondences and preserving style richness.

In the inference stage, we design an efficient region-based multi-style fusion module to embed in the middle of the backbone network. The module first segments the content feature map into regions based on semantic information, and then assigns the most suitable style to each region according to the correspondence confidences generated by the PA module. After decoding this hybrid future map, our network will create a seamless and coherent MST result.  Comprehensive evaluations show that our approach can produce more vivid and diverse results than existing SST and MST methods. 

In summary, the contributions in this paper are three folds: 

(1) We propose the first MST framework to automatically and spatially incorporate different styles into one result based on the semantic information. 

(2) We design a patch attention module for semantic correspondence, which broadens the form of attention module and enables the controllability of the size of the receptive field.
 
(3) We propose a conceptually simple yet effective region-based multi-style fusion module for MST to assign multiple styles to their semantically related regions and then seamlessly fuse them.


\begin{figure*}[htb]
    \centering
    \includegraphics[width=\linewidth]{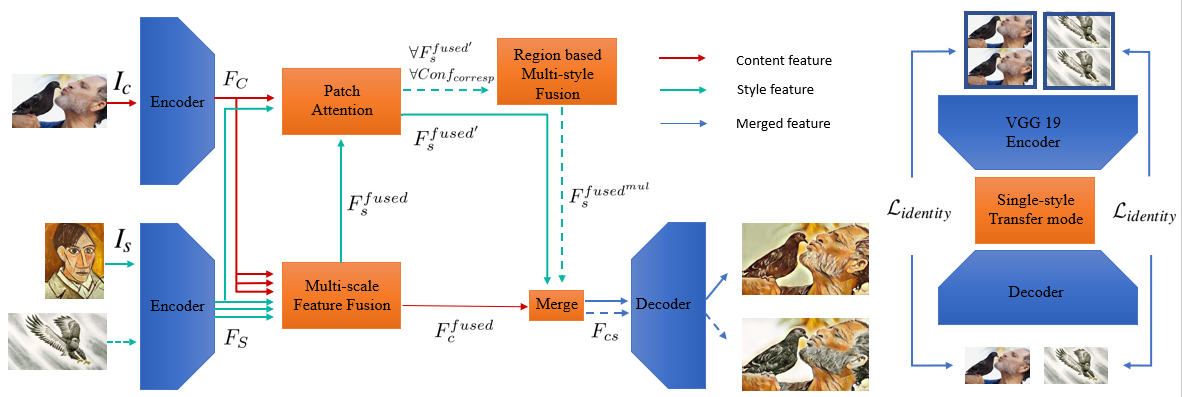}
    \caption{An overview of our proposed network.  Multi-level features $F_{C}$ and $F_{S}$  will first be fused in MFF module by channel-wise attention. Then the fused style feature $F_{s}^{fused}$ will be reassembled into $ F_{s}^{fused^{\prime}}$ guided by the semantic correspondence between $F_{c}^{relu4\_1}$ and $F_{s}^{relu4\_1}$. In SST case, $F_{s}^{fused}$ will be merged with $F_{c}^{fused}$ and decoded. While in MST case, as is shown by the dashed line, multiple $ F_{s}^{fused^{\prime}}$ and $conf_{corresp}$  will be fed to our multi-style fusion module and integrated based on regional correspondence confidence.}
    \label{fig:pipeline}
\end{figure*}
\section{Related Work}
\textbf{Neural style transfer.} Starting from the seminal work of Gatys \etal~\cite{gatys2015texture,gatys2016image}, Convolution Neural Network (CNN) demonstrates its remarkable ability to transfer style by matching statistical information between features of content and style images.
The framework of Gatys~\etal\cite{gatys2016image} is based on iterative updates of the image by optimizing content and style loss, which is applicable to arbitrary image but computationally expensive.
Numerous study have since been developed to improve style transfer in different aspects such as visual quality~ \cite{li2016combining,wang2017multimodal,risser2017stable}, perceptual control~ \cite{gatys2017controlling}, stroke control~ \cite{jing2018stroke,yao2019attention}.
A great number of researchers try to accelerate the transfer~ \cite{johnson2016perceptual,ulyanov2016texture, ulyanov2017improved,li2016precomputed,chen2017stylebank, li2017diversified,dumoulin2017learned} by approximating the iterative optimization with a feed-forward network. Although speed is improved dramatically, the flexibility is compromised since each network is restricted to a single style or a finite set of styles. The dilemma between speed, flexibility, and quality \cite{zhang2018metastyle} impedes the further development of style transfer.
Recently some fast Arbitrary-Style-Per-Model methods are proposed to resolve the dilemma. The idea is to train a style-agnostic autoencoder and convert the content feature into a given style domain while preserving content structures.  \cite{huang2017arbitrary,li2017universal,li2018learning} transfers the global style by coordinating the statistical distribution between them; while \cite{chen2016fast}  swap the content feature patch with the nearest style feature patch in terms of cosine similarity,  which achieves local semantic-aware style transfer results. Avatar-Net~ \cite{sheng2018avatar} further extends the AdaIN~\cite{huang2017arbitrary} to multi-scale style adaptation and loosen the restrictions of Style Swap \cite{chen2016fast} by performing projection before matching.  

Despite the success in SST, little attention has been paid to the field of MST, which is likely to create more vibrant and distinctive artistic effects. Some works extend their SST framework to MST as a simple add-in by linearly blending the feature from different styles \cite{gatys2016image,gu2018arbitrary,li2017universal,sheng2018avatar,park2018arbitrary,chen2016fast} or manually specifying the masks \cite{park2018arbitrary,li2017universal}. They either generate undesired results or require tedious user efforts. The challenge of MST is how to automatically combine the feature of different styles harmoniously without damaging the characteristics of each style. We effectively resolve this challenge by regional semantic matching and produce state-of-the-art MST results.

\textbf{Attention Module.}  Recently, attention mechanisms have become a key ingredient for models that need to incorporate global dependency~ \cite{bahdanau2014neural,gregor2015draw,xu2015show,yang2016stacked}. It allows the model to look globally but attend selectively at the data. Particularly, self-attention~\cite{hochreiter1997long,parikh2016decomposable} calculates the correlation between every two positions in a sequence. Such mechanism has been proved to be exceptionally effective in machine translation \cite{vaswani2017attention,bahdanau2014neural}, image classification \cite{xiao2015application,zhou2016learning}, visual question answering \cite{xu2015show} and image generation \cite{zhang2018self}.
Recently, \cite{park2018arbitrary} introduces style-attention to capture the correspondence between content image and style image and outperforms prior works in terms of visual quality. Compared to \cite{park2018arbitrary}, we further improve the capability of semantic matching to catalyze the performance of our multi-style fusion module.


\section{Proposed Method}
\label{sec:method}
The architecture of \emph{Style Mixer} is shown in Fig.~\ref{fig:pipeline}. The backbone style transfer model comprises an encoder and a decoder, with the multi-level feature fusion (MFF) module and patch attention (PA) module in the middle. In the case of MST, a multi-style fusion module is further embedded to distribute style features from different style references.
\subsection{Framework Pipeline} A pretrained VGG-19 network is employed as a feed-forward encoder to extract features of the input pairs. 
To incorporate multi-level feature produced by the encoder, an MFF module is placed after the encoder and takes features from 3 different layers as input.

Being able to classify the objects correctly despite the huge low-level variations, VGG-19 proves its efficiency and robustness in extracting semantic information. Therefore, by calculating the patch attention between the high-level feature of content and style images, we can obtain a meaningful semantic attention map and reassemble the style features accordingly. At last, we merge the reassembled style feature $ F_{s}^{fused^{\prime}}$ with $F_{c}^{fused}$ and decode them into an artistic image. 

Since the problems of multi-level feature fusion and semantic correspondence functions are common in both SST and MST, these two modules can be trained with SST and then applied to MST. In MST, Style Mixer will process multiple styles in a parallel manner, and incorporate them with our region-based style fusion strategy. The correspondence confidence $conf_{corresp}$ produced by PA module will guide the distribution of different styles based on semantic matching. In this way, every style will be assigned to the most semantically related region with local consistency. 

\subsection{Multi-level Feature Fusion Module}
\begin{figure}[htb]
    \centering
    \includegraphics[width=\linewidth]{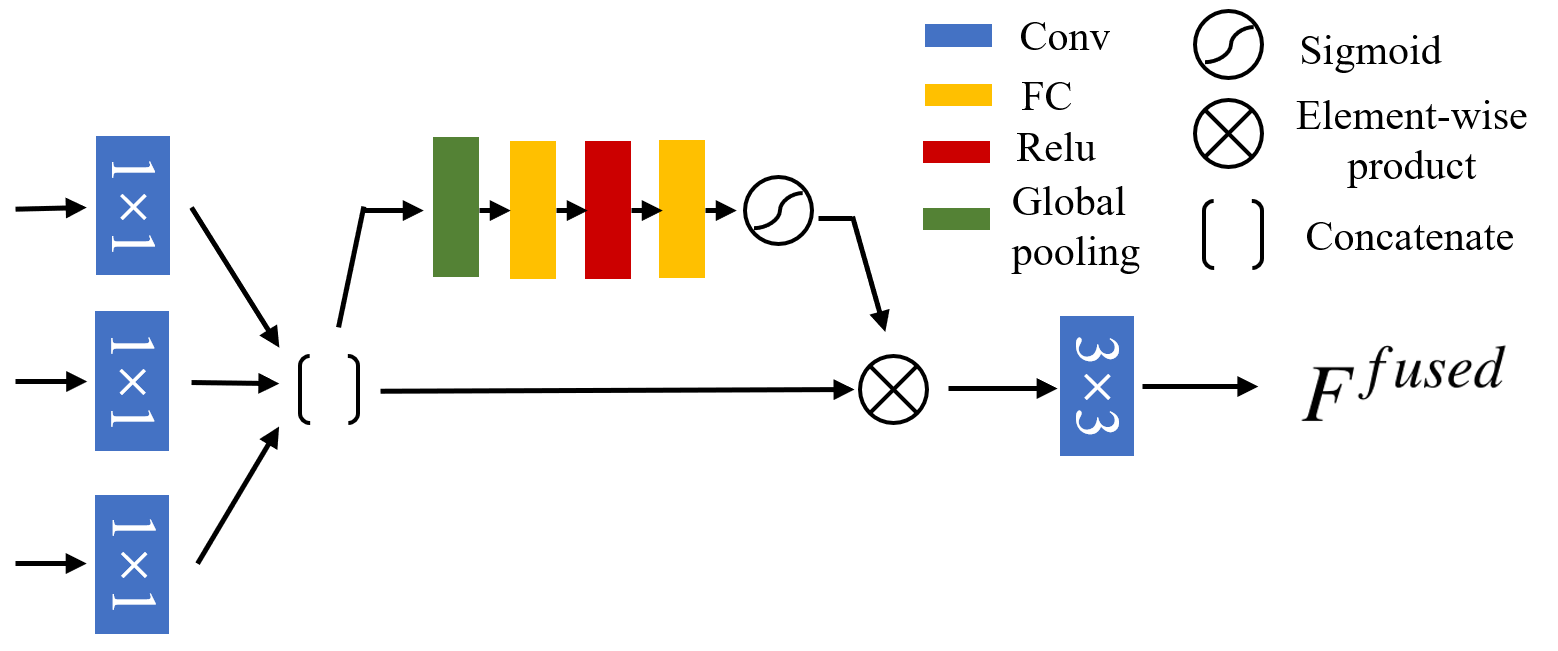}
    \caption{The process of multi-level feature fusion can be summarized as: concatenate, attend, and squeeze.}
    \label{fig:MFF}
\end{figure}
Features from different layers of VGG carry information of different scales and abstractness levels. To incorporate multi-level information, Avatar-net \cite{sheng2018avatar} introduces multi-level AdaIN \cite{huang2017arbitrary} to conduct style adaptation progressively. However, holistic statistic alignment sometimes creates unpleasant artifacts. After that, SANet \cite{park2018arbitrary} integrates two separate style-attention modules to extract style features of layer $relu4\_1$ and $relu5\_1$ to improve style richness but also introduces an expensive computational cost. To obtain faithful stylization with affordable computation cost (which is especially critical when adopting PA), we design an MFF module to coalesce the features from 3 different layers adaptively.

\begin{figure}[htb]
    \centering
    \includegraphics[width=\linewidth]{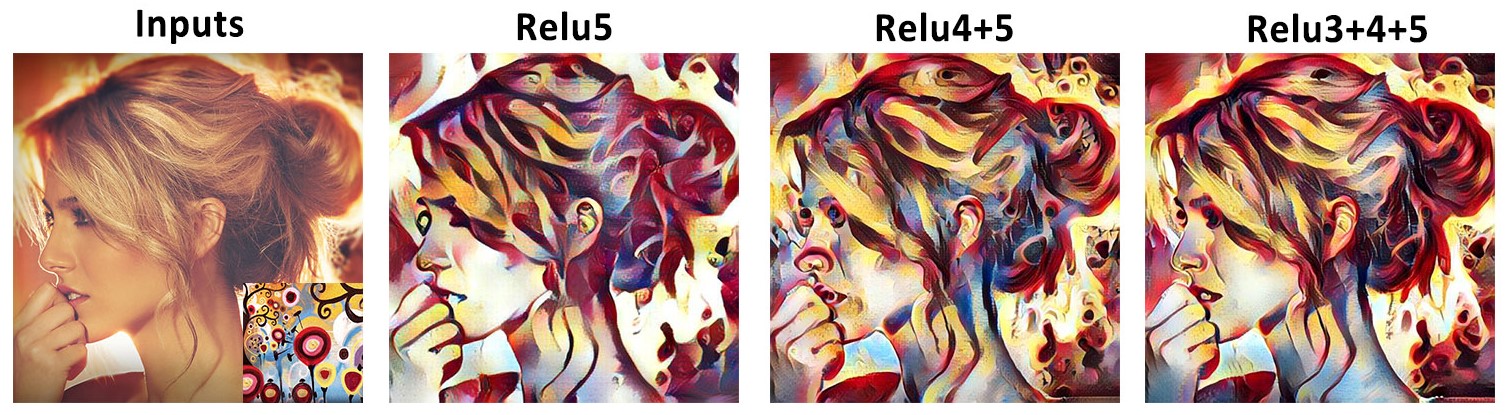}
    \caption{The second image, result of single-level feature, is rendered in large stroke and lacks style patterns. After adding the feature from $relu4\_1$, the result (the 3rd column) is richer in spiral patterns, for instance, the nose and upper-right corner. If we further integrate the feature from a lower level, which has a smaller receptive field, the high-frequency area, i.e., the hair and eyes of the women, become finer. At the same time, the cheek remains coarse. Features with different scales are combined pleasingly.}
    \label{fig:layers}
\end{figure}

The whole process of our MFF module is as depicted in Fig.~\ref{fig:MFF}. Features from $relu5\_1$, $relu4\_1$, $relu3\_1$ will first be recalibrated by a 1$\times$1 convolution. After that, all features will be resized to the same size and concatenated together.
In order to eliminate redundant and undesired feature, we conduct channel-wise attention~\cite{hu2018squeeze} to reweight the concatenated feature maps according to channel-wise importance.
At last, we apply one more 3$\times$3 convolution layer to smoothen the fused feature and obtain $F^{fused}$.
The comparison between different choices of input layers is shown in Fig.~\ref{fig:layers}

\subsection{Patch Attention Module}
\begin{figure}[htb]
    \centering
    \includegraphics[width=0.8\linewidth]{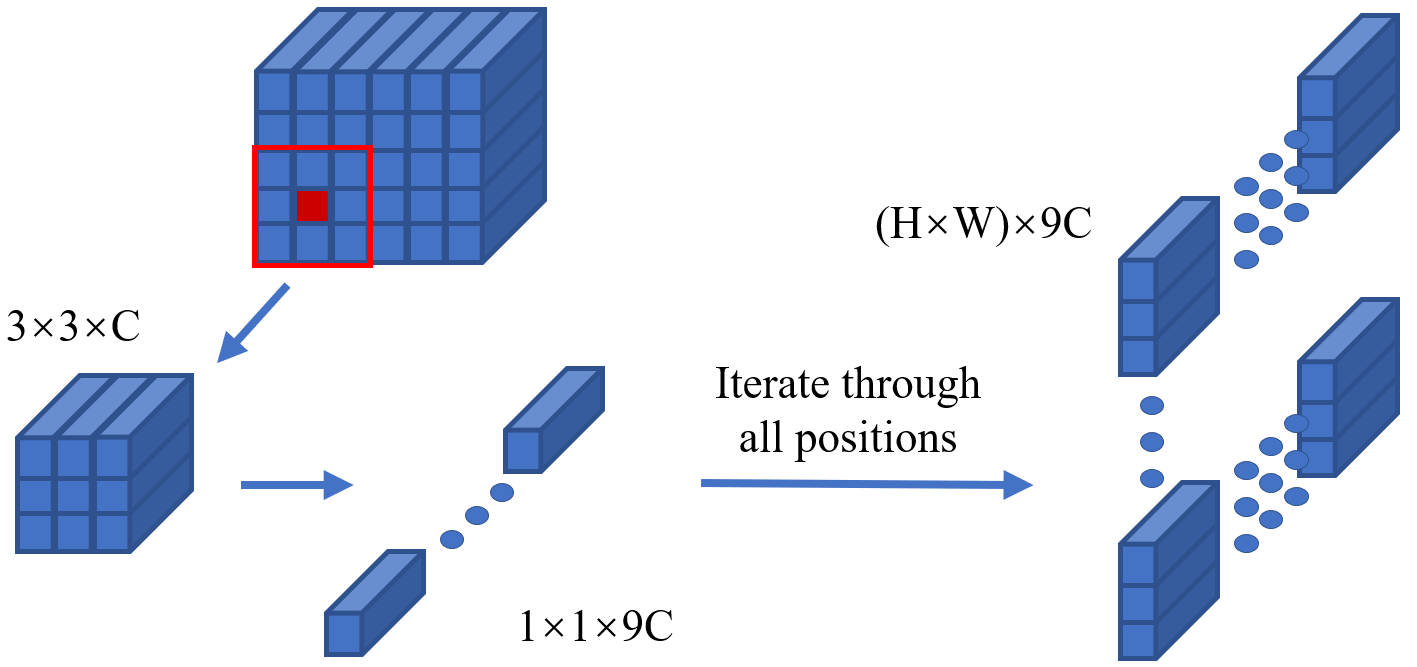}
    \caption{Unfolding operation.}
    \label{fig:unfold}
\end{figure}

\begin{figure}[htb]
    \centering
    \includegraphics[width=\linewidth]{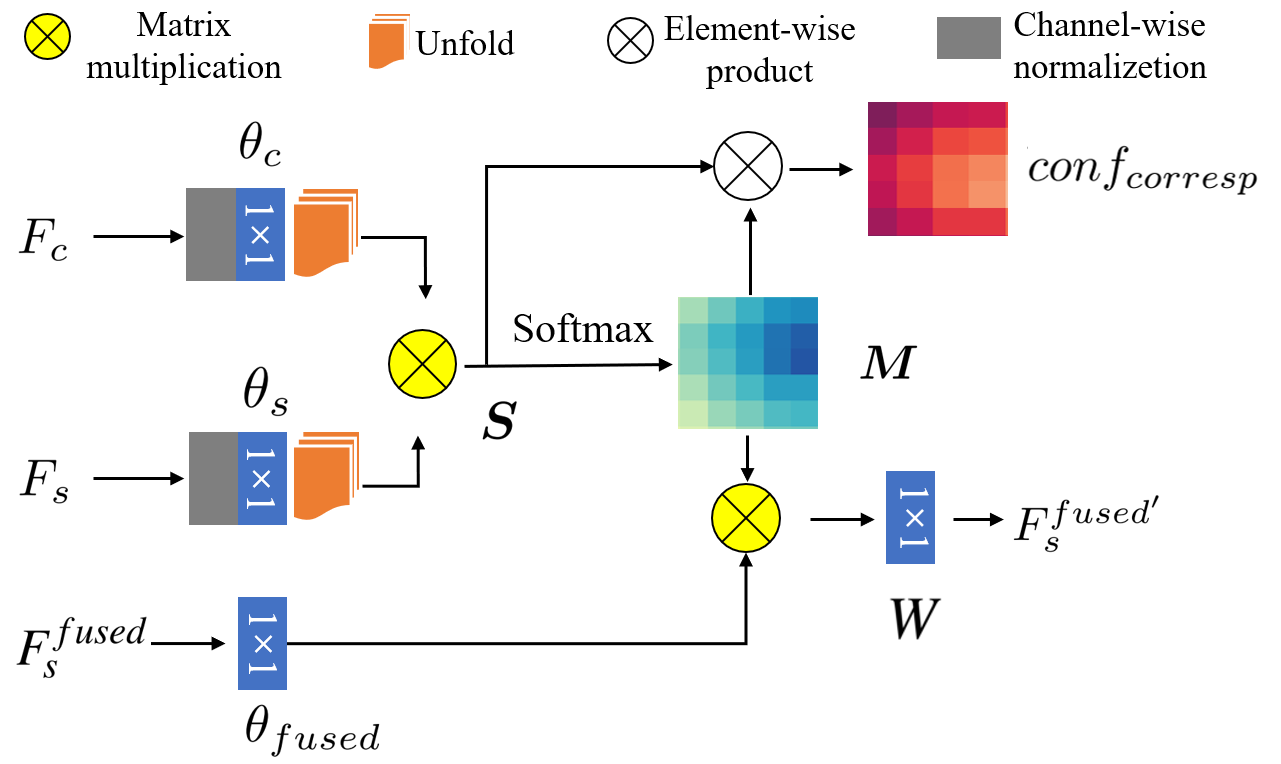}
    \caption{Patch attention module. In order to best preserve the semantic information, we calculate the correspondence score between original feature from VGG-19 to guide the reassembling of fused multi-level feature $F_{s}^{fused}$.}
    \label{fig:PA}
\end{figure}

Style Swap \cite{chen2016fast} is a pioneer work that introduces local patterns matching to style transfer. However, due to the fixed cosine similarity metric and the overlap between patches, it produces undesired overly smooth results with mismatches. SANet \cite{park2018arbitrary} proposed a novel style-attention mechanism to replace the fixed cosine similarity with a flexible learnable similarity kernel. Following the tradition of self-attention \cite{vaswani2017attention} and non-local block \cite{wang2018non}, it conducts point-wise attention between content and style features. Due to the limited size of the receptive field and local variation of the input image, point-wise attention performs unstably despite the learnable similarity kernel. To solve this problem, we extend the attention module to a more generic form, patch attention (PA), which enables the controllability of the size of the receptive field and better grasps the structural information. The mechanism of our PA module is illustrated in Fig.~\ref{fig:PA}. Together with the abundant semantic information in the high-level feature of VGG-19, our PA module achieves robust semantic matching. Also, it is worth noting that Style-attentional module in SANet is a special case of PA.

The PA module takes content feature $F_{c}$, style feature $F_{s}$ and $F_{s}^{fused}$ from MFF module as its inputs. It should be noted that in SANet \cite{park2018arbitrary}, attention is carried out between the content feature and the style feature, which will be reassembled. On the contrary, we calculated patch attention on the original feature of VGG-19, which is from layer relu\_4\_1, to best preserve the semantic information, and use the resulted pair-wise correspondence to guide the rearrangements of fused style feature $F_{s}^{fused}$. 
 PA starts with channel-wise normalization to put $F_{c}$ and $F_{s}$ into a common domain. This can be regarded as style normalization \cite{huang2017arbitrary,li2017demystifying} and encourages matching to rely only on structural and semantic similarities. Then we perform a $1\times1$ convolution to the normalized feature to enable the learning of a suitable similarity kernel by itself. To improve the matching accuracy, we take neighboring information into consideration by unfolding patches at each position. The unfold operation is demonstrated in Fig.~\ref{fig:unfold}. In Eq.~\ref{eq:nonlocal1}, $\overline{F}$ represents channel-wise normalized feature, and $P^i$ indicates a vectorized patch feature at $i$-th position, which consists of the information of the $i$-th position and its neighborhood.
\begin{equation}
\begin{split}
P_{k}^i=Unfold(\theta_{k}(\overline{F_{k}}))^i,\ \ where\ \ k \in \{{s},{c}\}.
\end{split}  
\label{eq:nonlocal1}
\end{equation}

Next, the correspondence score $\boldsymbol{S}$ and semantic attention map $\boldsymbol{M}$ are calculated with patch attention mechanism as Eq.~\ref{eq:nonlocal2}. After performing softmax operation on each row of $\boldsymbol{S}$, we obtain the attention map needed for the reallocation of $F_{s}^{fused}$:
\begin{equation}
\begin{split} 
M_{i,j}=\frac{\exp(S_{ij})}{\sum_{j=1}^{N}\exp(S_{ij})},\ \ where\ \ S_{ij}=({P}_{c}^{i})^T{P}_{s}^j\\
and\ \ N=spatial\ \ size\ \ of\ \ F_{s}.
\end{split} 
\label{eq:nonlocal2}
\end{equation}
Driven by contextual loss and identity loss, similar features will obtain a larger correspondence score, resulting in the larger attention value in the $\boldsymbol{M}$. Thanks to the rich semantic information provided by encoder, the correspondence score can be interpreted as semantic affinity. Thus, in the reallocation process, as is depicted in Eq.~\ref{eq:nonlocal3}, style feature that is more semantically related will be emphasized. $F_{s}^{fused^{\prime}}$ refers to the reassembled style feature from PA module.
\begin{equation}
\begin{split} 
F_{s}^{fused^{\prime}}=M\theta_{fused}(F_{s}^{fused}).
\end{split} 
\label{eq:nonlocal3}
\end{equation}

To measure the confidence that $F_{s}^{fused^{\prime}}$ have same semantic implication as $F_{c}$, we further conduct element-wise multiplication between correspondence score $\boldsymbol{S}$ and semantic attention map $\boldsymbol{M}$ to derive a correspondence confidence $conf_{corresp}$.
In essence, $conf_{corresp}$ is the weighted average correspondence score of $F_{s}^{fused^{\prime}}$, representing the semantic correspondence between a given style feature $F_{s}$ and $F_{c}$. $conf_{corresp}$ plays a critical role in the distribution of styles in MST. We define it as:
\begin{equation}
\begin{split} 
conf_{corresp}^{i}=\sum_{j=1}^{N}S_{i,j}M_{i,j},
\end{split} 
\label{eq:CFM1}
\end{equation}
where $conf_{corresp}^{i}$ indicates the correspondence confidence of $F_{s}^{fused^{\prime}}$ at location i.
\begin{figure}[htb]
    \centering
    \includegraphics[width=\linewidth]{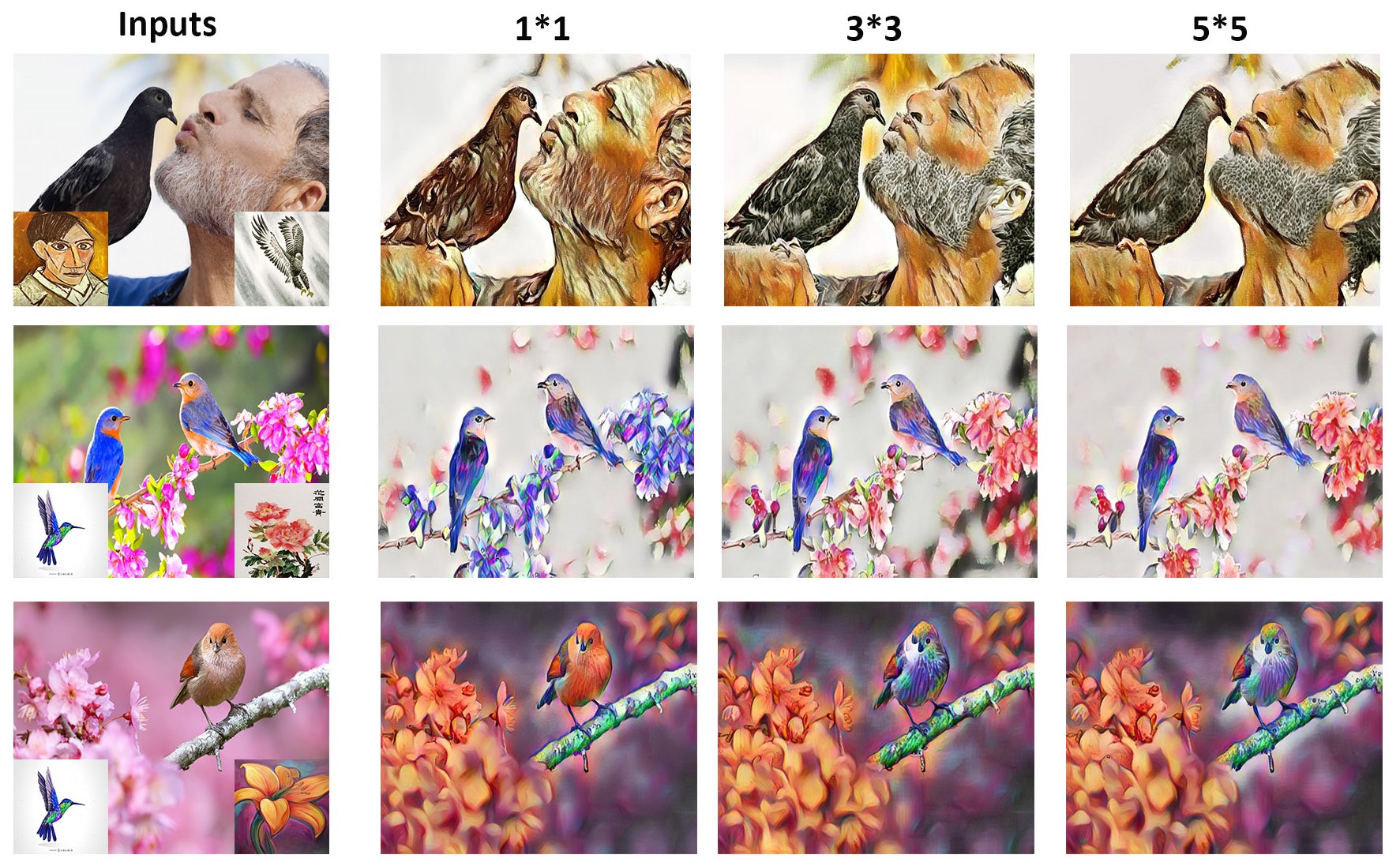}
    \caption{
    Investigation of patch size. $1\times1$ PA completely fails to differentiate bird with people or flower. While $5\times5$ PA obtains a good overall matching accuracy, it sometimes mismatches objects due to noisy neighboring information, i.e., some flowers in the second example are mistakenly identified as background and disappear.
    }
    \label{fig:patch}
\end{figure}

The size of the receptive field is an intrinsic characteristic of a chosen layer and always fixed. PA enables the adjustability of the receptive field and further releases the potential of attention mechanism. From Fig.~\ref{fig:patch}, we may see how different patch size affect matching and stylization results. In all 3 cases, $1\times1$ (traditional point-wise attention) failed to capture semantic correspondence correctly. In the first pair, the bird was wrongly rendered in the style of the portrait. While in the other two pairs, styles of bird and flower respectively dominate the whole image, disregarding the semantic meaning of different objects. On the contrary, both $3\times3$ and $5\times5$ PA demonstrate an excellent capability of semantic matching. However, larger patch size tends to compromise the detail. For instance, in the third image of the second row, some flowers in the background disappear. It is probably because the neighboring information dominates the matching so that the flowers wrongly match with the background of the styles. In addition, with consideration of computation cost in mind, we choose $3\times3$ PA in our model.

\subsection{Region-based Multi-style Fusion Module}
\begin{figure*}[htb]
    \centering
    \includegraphics[width=\linewidth]{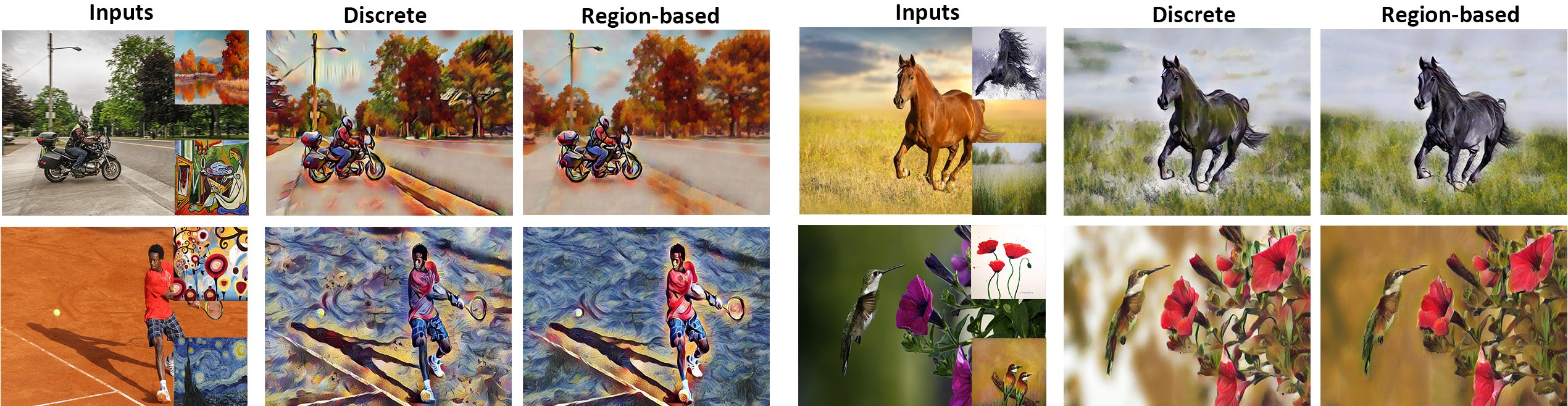}
    \caption{Comparison between region-based strategy and discrete strategy. The discrete strategy introduces noises in certain regions, eliminates the characteristics of style features and therefore fails to generate high-quality stylized images.}
    \label{fig:discrete}
\end{figure*}
In MST, the most challenging problem lies in how to harmoniously incorporate different styles without hurting the characteristics of each style. This has two underlying implications. 

Firstly, styles should not be mixed; otherwise, they will obfuscate each other and compromise style integrity. What is worse, mixing distinctive styles may produce disturbing and nondescript patterns. Thus, the assignment of different styles should be mutually exclusive.
Secondly, a metric needs to be defined to decide the distribution of multiple styles. Semantic correspondence is a natural idea since, with semantic consideration, the overall effect will look more reasonable and intuitive. Correspondence confidence $conf_{corresp}$ is precisely the objective measure of semantic correspondence among different styles.

Given the two consideration above, a straightforward idea is to assign the style with the highest confidence to each position. However, local variation and noise sometimes intervene in the calculation of correspondence, inducing false match, and producing unpleasing discrete patterns. In Fig.~\ref{fig:discrete}, we can see that the discrete strategy produces many scattered pattern and deteriorates local consistency. 

To resolve the problem, we utilize clustering to segment our content feature map ($relu4\_1$) and calculate regional correspondence confidence. The regional voting strategy increases the robustness of matching by fixing individual mismatch. As we mentioned before, high-level feature comprises abundant semantic information, clustering in high-dimensional feature space is efficient in distinguish objects with different semantic implication. Specifically, we apply K-means to cluster all feature vectors as well as their spatial location in Euclidean Distance to ensure spatial affinity of the result.

The pipeline of MST is depicted as the dashed line in Fig.~\ref{fig:pipeline}. MFF module and PA module will process multiple style references in a parallel way and pass all the reassembled style features $F_{s}^{fused^{\prime}}$ and correspondence confidence $conf_{corresp}$ to multi-style fusion module.

To allocate semantically nearest style for each region, we calculate the regional sum of correspondence confidence and choose the style with the highest value for each region. The assignment policy is conceptually simple but proves its robustness by comprehensive evaluation. Formally, let $R$ to be a specific region, we calculate the sum of correspondence confidence in R for every style, and style $k$ with the highest sum will be the assignment result $I_{R}$ for region R. Formally, the strategy is defined as:
\begin{equation}
\begin{split} 
I_{R}=\mathop{argmax}\limits_{k}(\sum_{i\in R}conf_{corresp}^{i,k}),
\end{split}
\label{eq:CFM2}
\end{equation}
where $conf_{corresp}^{i,k}$ indicates the correspondence confidence of style $k$ at position $i$. 

Compared to the straightforward discrete strategy, our proposed region-based strategy improves the visual quality and matching robustness. In Fig.~\ref{fig:discrete}, the results of discrete strategy are suffered from mismatch and local inconsistency, such as the blemishes on the horse and grassland in upper-right pair. By conducting regional voting, those flaws are fixed automatically. Both horse and grass are faithfully transformed according to the reference image.
\begin{figure}[htb]
    \centering
    \includegraphics[width=\linewidth]{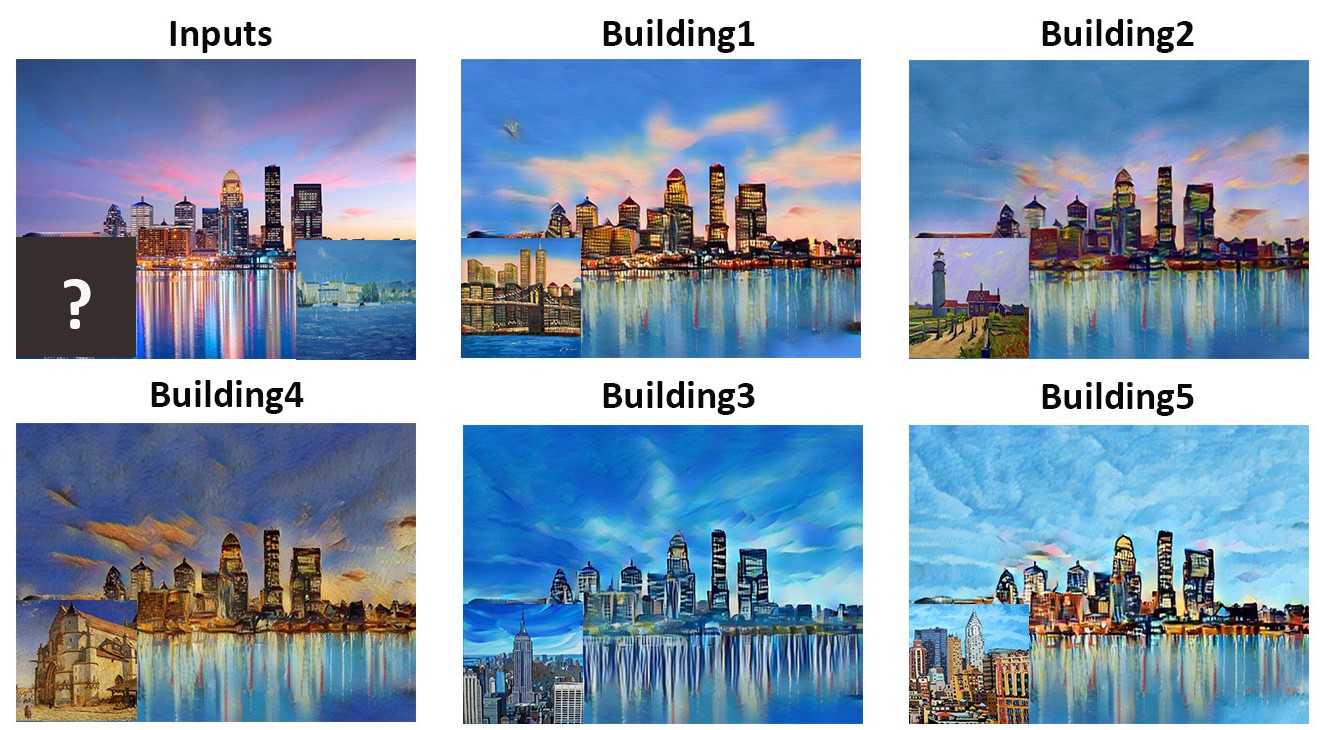}
    \caption{By changing different styles of architecture, the corresponding region in content image changes simultaneously.}
    \label{fig:control}
\end{figure}
With Fig.~\ref{fig:control}, you will have a better idea about how the styles are distributed.
\begin{figure*}[htb] 
    \centering
    \includegraphics[width=\linewidth]{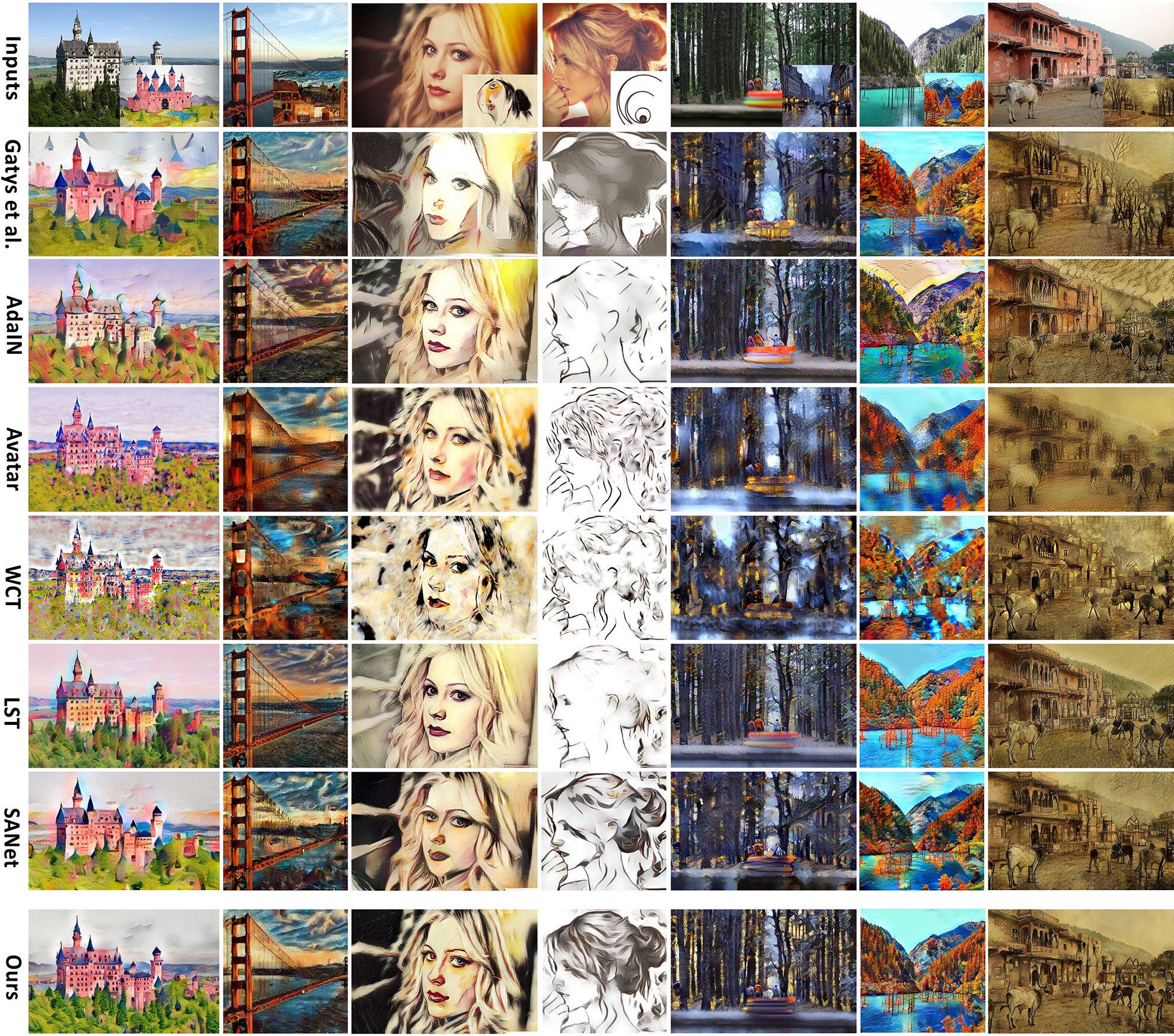}
    \caption{Visual comparison with existing works on SST.}
    \label{fig:SST}
\end{figure*}
\section{Experiments and Results}
\subsection{Implementation Details}
We train our network using MSCOCO and WikiArt datasets as content images and style images, respectively, both of which contain roughly 80000 images. We use an Adam optimizer to train the backbone model with a batch-size of 6 content-style pairs and a learning rate initially set to $1e-4$. During the training process, we firstly resize the smaller dimension to 512 pixels while preserving the aspect ratio, and then randomly crop regions of $256\times256$ pixels for end-to-end training.

Our loss function is defined as below to drive the training process:
\begin{equation}
\begin{split} 
\mathcal{L}=\lambda_{c}\mathcal{L}_{c}+\lambda_{s}\mathcal{L}_{s}+\mathcal{L}_{identity}+\lambda_{cx}\mathcal{L}_{cx}.
\end{split}
\label{eq:LOSS1}
\end{equation}

Similar to \cite{huang2017arbitrary}, our perceptual loss $\mathcal{L}_{c}$ is defined as Euclidean distance between channel-wise normalized VGG-19 features extracted from content image and synthesized image. Feature layer $relu3\_1$, $relu4\_1$ and $relu5\_1$ are used to compute perceptual loss. For style loss $\mathcal{L}_{s}$, we apply style loss same as AdaIN \cite{huang2017arbitrary} to drive the global style transfer.

We also apply contextual loss proposed by \cite{mechrez2018contextual} to facilitate the semantic matching between style feature and content feature. The cosine distances $d_{i,j}^L$ are calculated between each pair of feature vectors in the feature maps of style and synthesized image. After $d_{i,j}^L$ being normalized as $\overline{d^L(i,j)}=\frac{d_{i,j}^L}{\min \limits_{k}d_{i,k}^L+\epsilon}$, the affinity between any two feature points in layer $L$ is represented as: 
$$A^L(i,j)=\mathop{softmax}\limits_{j}(1-\overline{d^L(i,j)}/bw),$$
where $bw$ is the bandwidth, typically set to 0.1.
The contextual loss is defined to maximize such affinity between the synthesized image and the semantically nearest style feature:
\begin{equation}
\begin{split}
\mathcal{L}_{cx}=\sum_{L}[-log(\frac{1}{N_{L}}\sum_{i}\max_{j}A^L(i,j))],
\end{split}
\label{eq:LOSS4}
\end{equation}
where $N_{L}$ is the number of feature vectors at layer L and l is set to $2$ to $4$ in our case.
\begin{figure*}[htb]
    \centering
    \includegraphics[width=\linewidth]{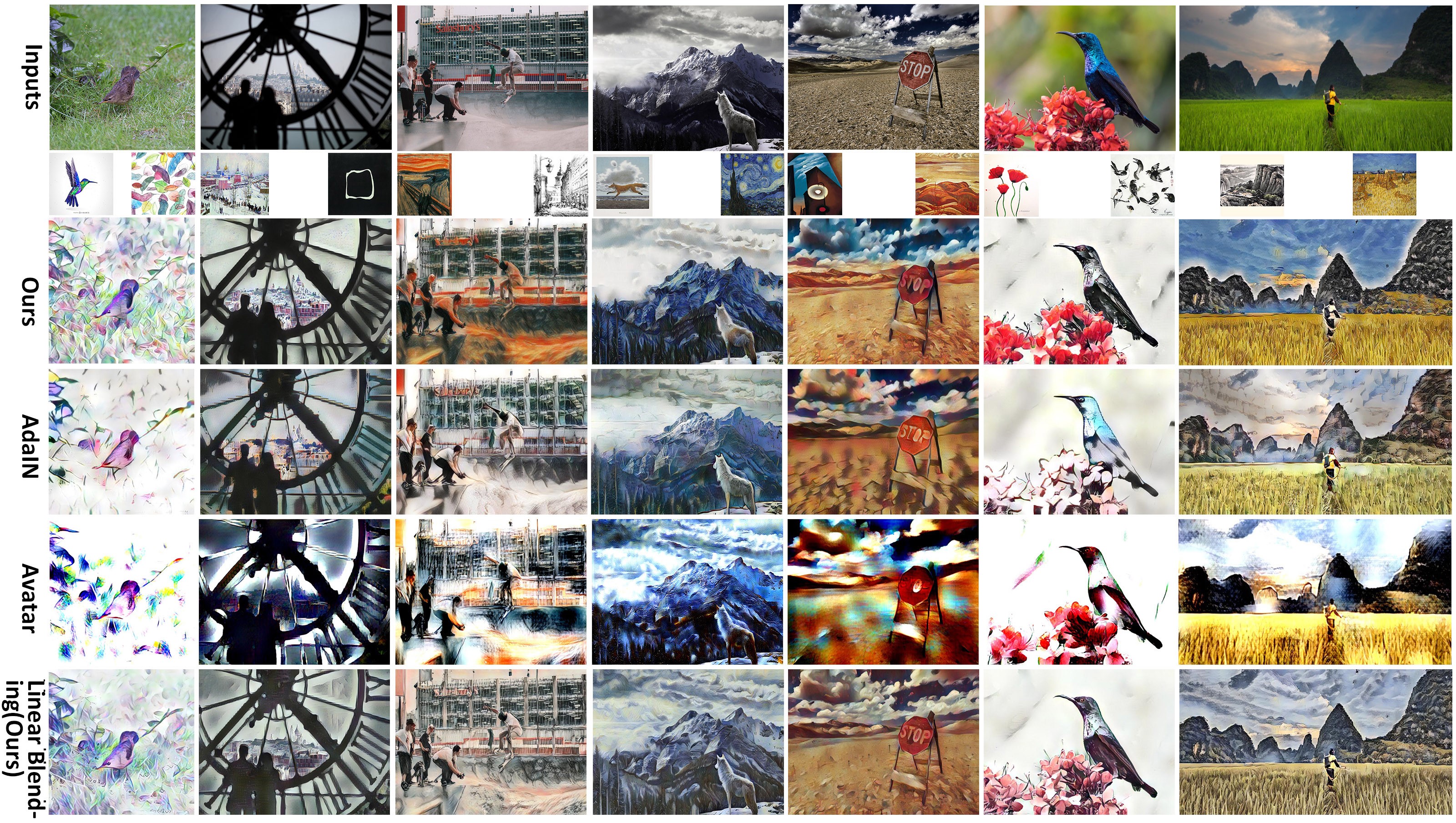}
    \caption{Visual Comparison with existing works on MST.}
    \label{fig:MST}
\end{figure*}

In order to guide the network to gain the powerful ability of semantic matching and image reconstruction, an advanced identity loss proposed by \cite{park2018arbitrary} is employed, as is shown in Fig.~\ref{fig:pipeline}. Two symmetric pairs of content and style images are fed to the network with the hope that the network should be able to reconstruct the original images, and the results are identified as $I_{cc}$ and $I_{ss}$ separately. Formally, the identity loss is defined as below:
\begin{equation}
\begin{split} 
\mathcal{L}_{identity}=\lambda_{identity1}(||({I}_{cc}-{I}_{c})||_{2}+||({I}_{ss}-{I}_{s})||_{2})
\\+\lambda_{identity2}\sum_{i=1}^5(||{VGG}_{i}({I}_{cc})-{VGG}_{i}({I}_{c})||_{2}
\\+||{VGG}_{i}({I}_{ss})-{VGG}_{i}({I}_{s})||_{2}).
\end{split}
\label{eq:LOSS5} 
\end{equation}
In addition, we change the behavior of merging module during identity loss calculation to:
\begin{equation}
\begin{split} 
F_{cs}=k\times F_{s}^{fused\prime},
\end{split}
\label{eq:amp} 
\end{equation}
where k is a learnable scale factor and we name the module as Amplifier. The advantage of Amplifier is further discussed in sec.~\ref{sec:dis_amp}.

The weight parameters $\lambda_{c}$, $\Lambda_{s}$, $\lambda_{cx}$, $\lambda_{identity}$, $\lambda_{identity2}$ are set to 3, 3, 3, 1, 50 respectively according to our experiments.
\label{sec:experiment}

\subsection{Qualitative Comparison}
\begin{figure*}[htb]
    \centering
    \includegraphics[width=\linewidth]{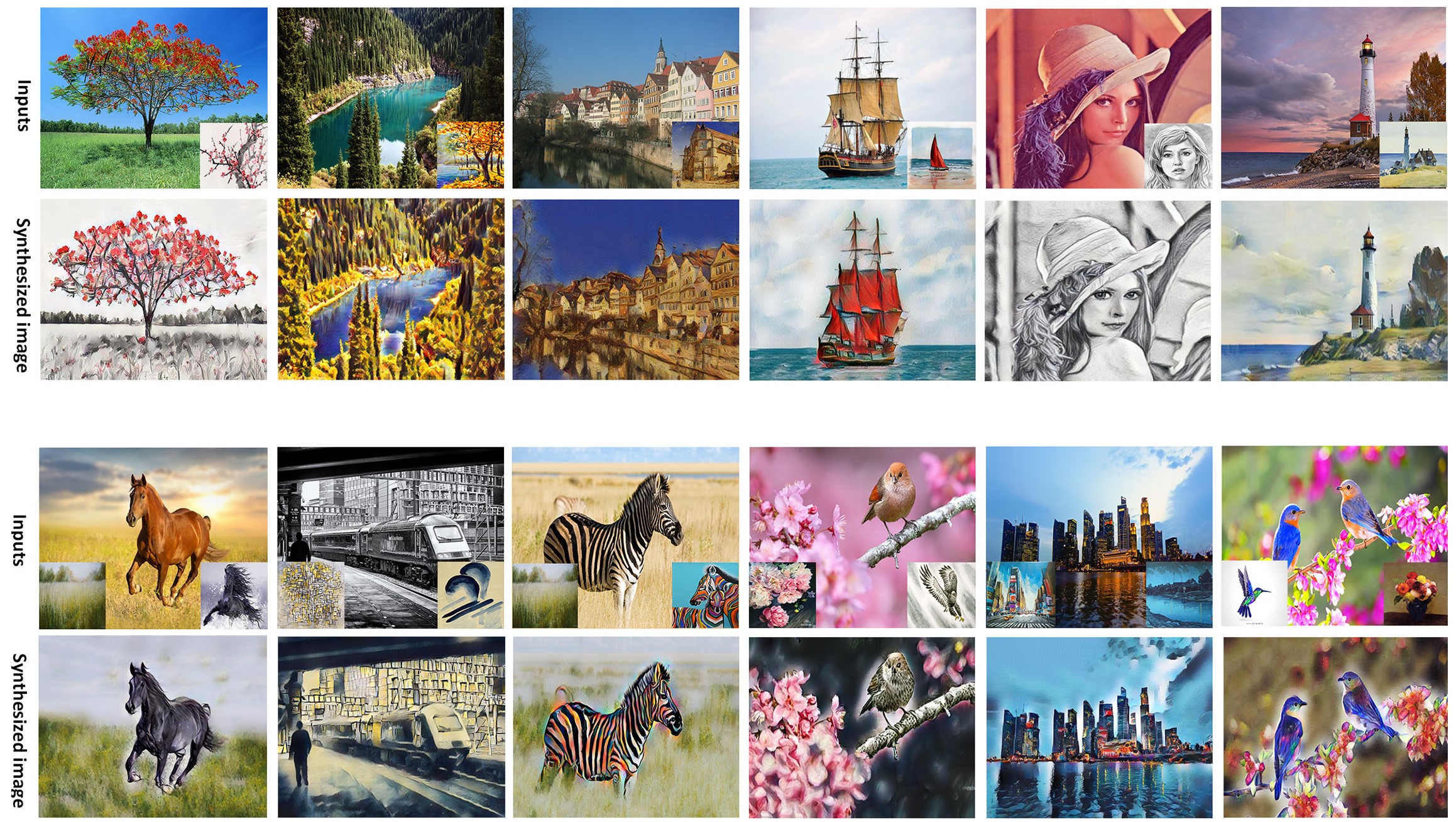}
      \caption{More results of Style Mixer. The upper rows are results of SST which showcase the competency of semantic matching of our backbone model, i.e., the eyes of the lady are transferred accordingly in the 5th column. The two rows below provide more examples of results to demonstrate the superiority of Style Mixer in terms of MST.}
    \label{fig:appendix-multi2}
\end{figure*}

 To evaluate the effectiveness of our backbone model and region-based style fusion strategy, we conduct a comparison with existing methods. All the inputs are chosen outside the training set. For a fair comparison, we generate results by running the released codes of the aforementioned works with the default configuration, except for SANet (We use the official \href{http://style.airi.kr/ori\_demo/}{demo page}). The visual comparisons of SST and MST methods are shown in Fig.~\ref{fig:SST} and Fig.~\ref{fig:MST} respectively. Additionally, extra examples of our work can be found in Fig.~\ref{fig:appendix-multi2}. 

\textbf{Single-style transfer.} Single style performance comparison results are available in  Fig.~\ref{fig:SST}. The optimization-based method \cite{gatys2016image} is unstable since it is likely to stick in the local minimum for some pairs, which can be seen in column 3, 4 of Gatys et al. in Fig.~\ref{fig:SST}. The two faces suffer heavily from the loss of details and deviation of style. Both AdaIN \cite{huang2017arbitrary} and WCT \cite{li2017universal} holistically adjust the content features to match the global statistics of the style features, which leads to blurring effect and textual distortion in some local regions (e.g., the last column of AdaIN and WCT, the pattern of trees grow indiscriminately to the sky). Although Avatar Net \cite{sheng2018avatar} shrinks the domain gap between content and style features and utilizes patch-wise semantics, it tends to produce fuzzy effects due to overlapping patches and repeated patterns because of global statistical alignment (e.g., column 1, 2, 7 of Avatar in Fig.~\ref{fig:SST}). LST \cite{li2018learning} originates from \cite{li2017universal} and generates some good results, but it is vulnerable to wash-out artifacts (e.g., column 3, 4 of LST in Fig.~\ref{fig:SST}) and halation around the edges (e.g., column 1, 6). Besides, this method fails to display desired stylized effect for some images (e.g., column 2, 5 of LST). SANet \cite{park2018arbitrary} applies style-attention mechanism to flexibly conduct style transfer. However, false matching and distortions still occur for this method, such as the pink pattern on trees in the first column.

Our method achieves the most balanced performance among all the above models. Our method greatly improves the content preservation by incorporating content features from relu3, 4, 5, which can be seen in column 1, 5, 6, 7 of Fig.~\ref{fig:SST}. At the same time, it presents rich style patterns that are both appealing and meaningful (e.g., column 2, 4 of Ours in Fig.~\ref{fig:SST}). Besides, learnable patch attention module takes contextual information into consideration and flexibly reassembles style patterns, which makes a breakthrough in semantic feature transfer (e.g., column 1, 3, 6).

\textbf{Multi-style Transfer.} To illustrate the effectiveness of our region-based strategy for MST, we compare it with the traditional linear blending strategy implemented by AdaIN \cite{huang2017arbitrary}, AvatarNet \cite{sheng2018avatar} as well as our backbone model.

All the results are shown in Fig.~\ref{fig:MST}. Generally speaking, linear blending mixes different styles; therefore, the characteristics of the individual style are not preserved. It tends to produce muddled results with fade-out effects. By applying linear blending strategy, our model and AdaIN \cite{huang2017arbitrary} fail to retain characteristics of individual style as the structural and color information is fused indiscriminately (column 2, 3, 5, 7 in Fig.~\ref{fig:MST}).
Although AvatarNet \cite{sheng2018avatar} preserves the style patterns for certain images, it seriously suffers from fade-out effects (column 2, 5, 6, 7 of Avatar in Fig.~\ref{fig:MST}). On the other hand, Style Mixer eliminates the interference between different styles with a spatially exclusive transfer strategy. In the last column of Fig.~\ref{fig:MST}, three linear blending based methods produce results with colors that do not exist in style references, while our proposed Style Mixer faithfully transfer the field, mountain, and sky in style references to the result.
\subsection{Quantitative Comparison}
\begin{figure}[htb]
    \centering
    \includegraphics[width=\linewidth]{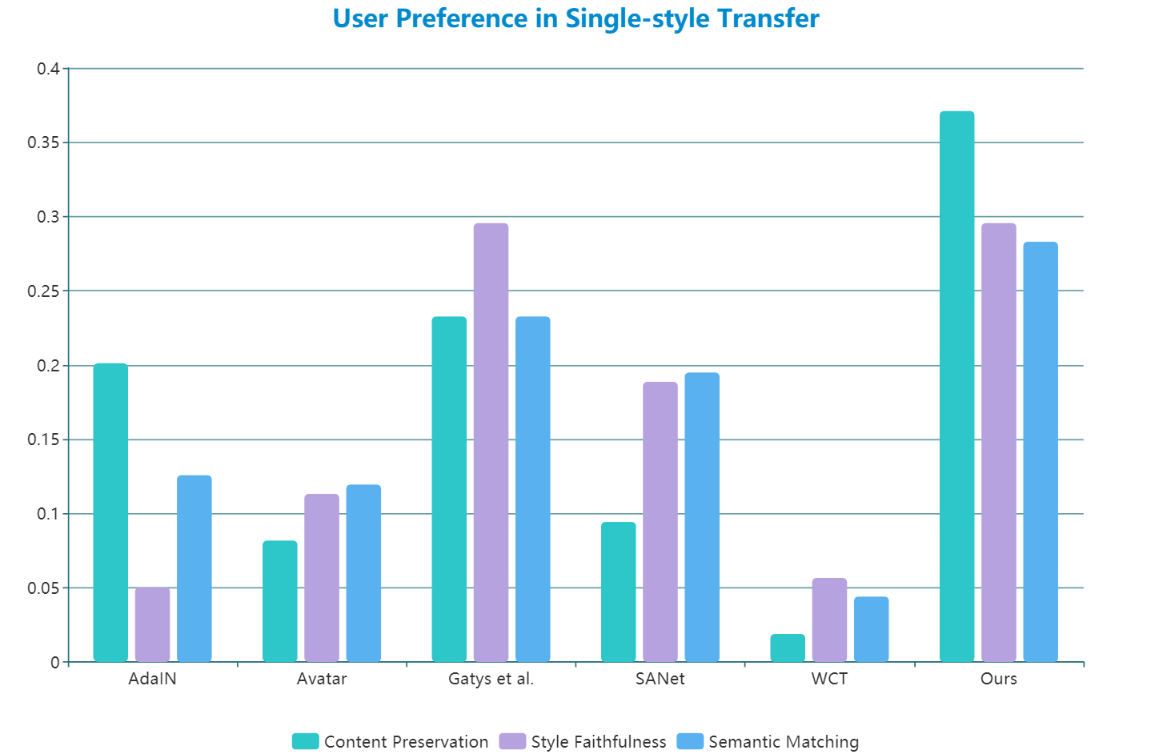}
    \caption{User preference towards different SST algorithms in terms of different metrics.}
    \label{fig:us1}
\end{figure}
\begin{figure}[htb]
    \centering
    \includegraphics[width=1\linewidth]{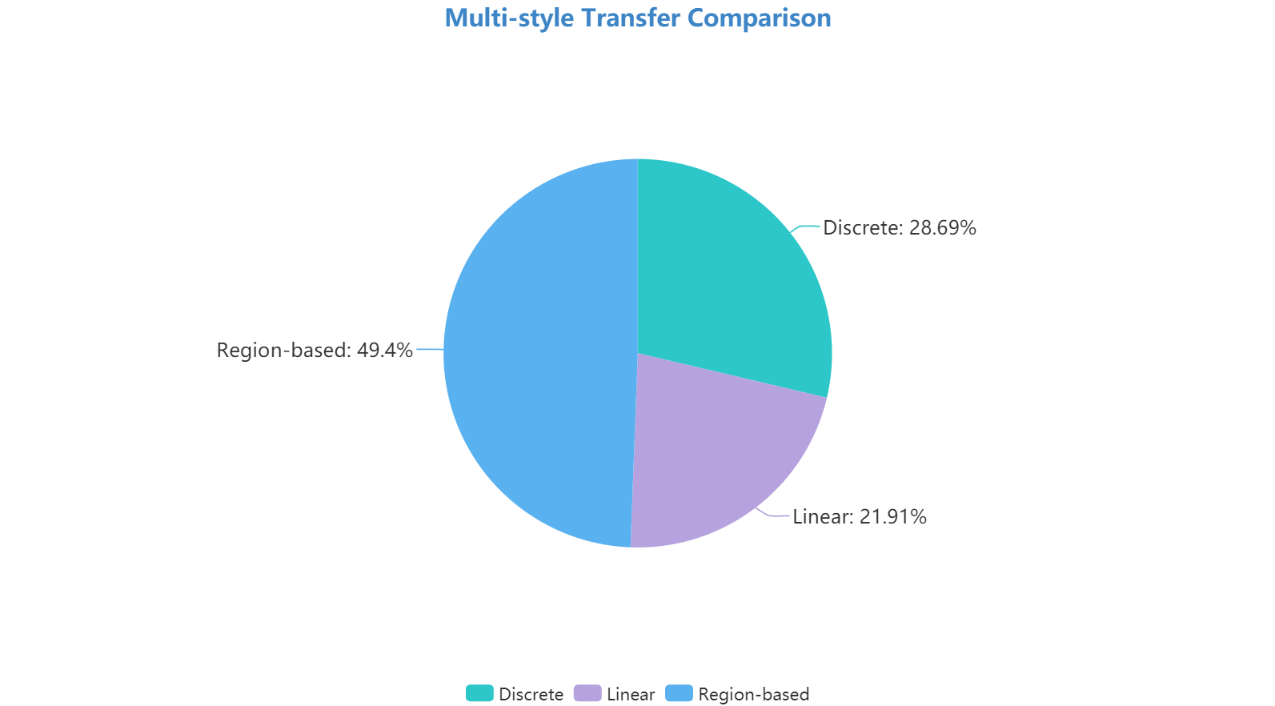}
    \caption{User preference towards different MST strategies.}
    \label{fig:us2}
\end{figure}
In order to validate our work, we further conduct two user studies to evaluate the SST performance of our backbone model and MST performance of Style Mixer. Both studies are conducted among 40 participants uniformly ranging from university students to normal officers. For each question, we display the results of all methods in random order and ask the participants to choose the one that best conforms to the given metrics. All the questions are presented in random order, and the participants are given unlimited time to finish the questions.
Unlike the settings in regular user studies, we do not choose the test images randomly. Instead, we handpicked semantically related content and style image pairs to evaluate the performance on semantic matching. Each user studies involves 36 pairs of images in total, and each user will be presented with six randomly chosen ones.
\begin{figure}[htb]
    \centering
    \includegraphics[width=1\linewidth]{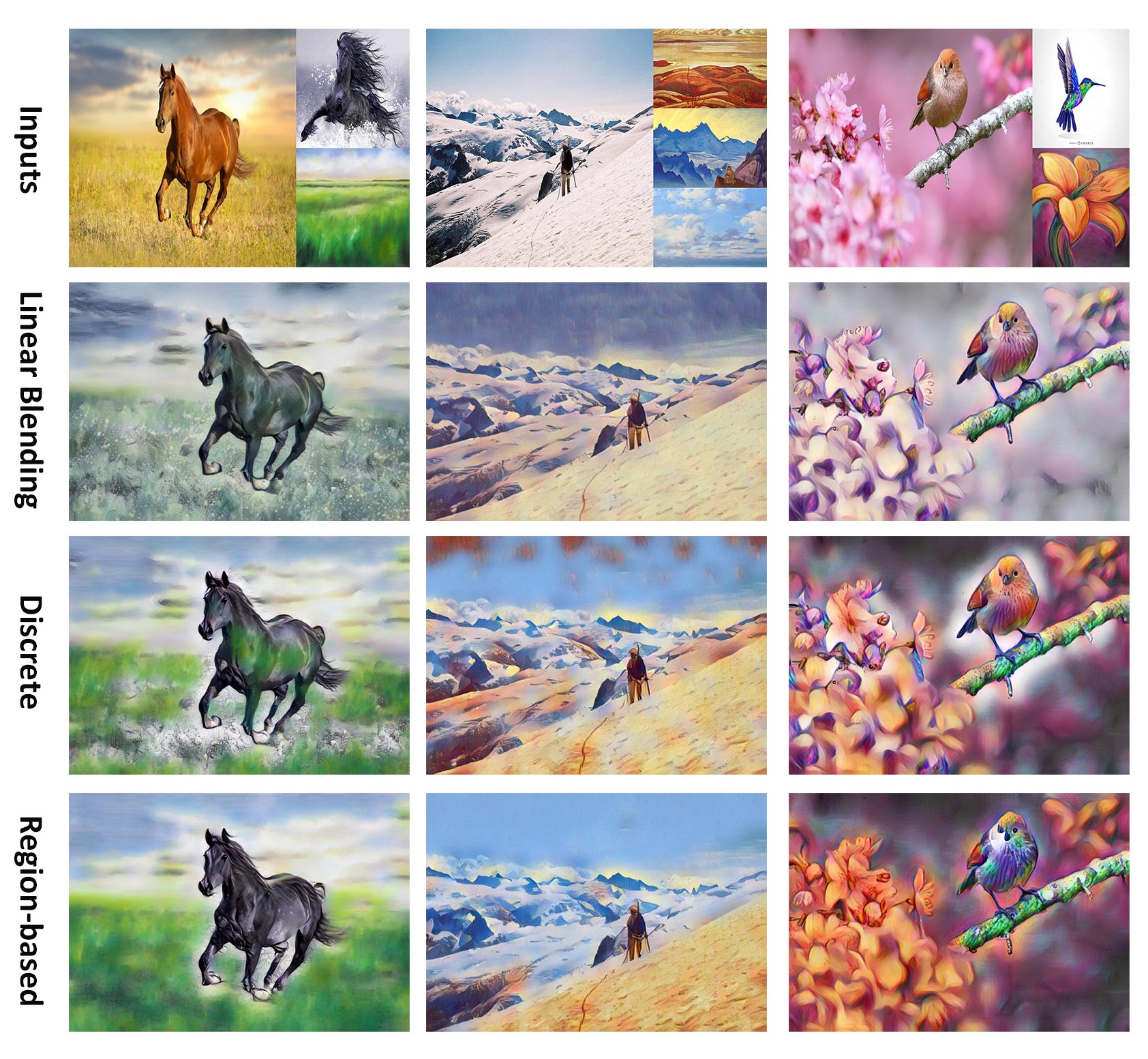}
    \caption{Exemplar images in MST survey. The results of our region-based fusing strategy demonstrate the best style faithfulness with local consistency.}
    \label{fig:us2_sample}
\end{figure}

\textbf{Single-style Transfer.} Firstly, we access the ability of our backbone model on SST. 5 state-of-the-art models \cite{gatys2016image,huang2017arbitrary,li2017universal,sheng2018avatar,park2018arbitrary} are chosen for comparison. We follow \cite{yao2019attention} to evaluate content preservation and style faithfulness. Besides, we introduce the semantic matching ability as a new metric, indicating whether the styles are transferred according to semantic matching, i.e., tree-to-tree, face-to-face. We manually make explicit instructions with exemplar images to define the criteria for each metric. For a fair comparison, we run the released code with the default setting for the aforementioned models. As we can see in Fig.~\ref{fig:us1}, our model obtains the most impressive performance in visual perspective, especially in content preservation. Even in terms of style faithfulness, our model is competitive with iteration-based method \cite{gatys2016image}. Also, the semantic matching score of our proposed method is the highest among the six models, and this should be credited to the PA module. The extraordinary visual quality and semantic matching of our backbone model serve as the cornerstone of our MST framework.

\textbf{Multi-style transfer.} In order to evaluate the user preference towards different MST strategies, we eliminate the effect of the backbone model by using the same one (our model) for all strategies. Our region-based strategy is compared with linear blending as well as the discrete strategy in the user study.

The result illustrated in Fig.~\ref{fig:us2} shows that our region-based strategy is superior to the other two methods. Linear blending is the least favorable probably because of the muddled results and insipid color, as is shown in \ref{fig:us2_sample}. The discrete strategy produces more vivid results with some flaws due to unstable local matching (i.e., the green color on the horse in the first image and mottled sky in the second image of Fig.~\ref{fig:us2_sample}). While our proposed method fixes those false matching by regional voting mechanism and thus obtains more decent results.
\begin{table}[]
\begin {center}
\begin{tabular}{c|cc}
Method & SST Time  & MST Time\\ 
\hline
 Gatys et al. \cite{gatys2016image} & 51.04 & -\\
 AdaIN \cite{huang2017arbitrary} & 0.014 & 0.032 (Linear blending) \\
 WCT \cite{li2017universal} & 0.933 & - \\
 Avatar-Net \cite{sheng2018avatar} & 0.330 & 0.526 (Linear blending)  \\
 SANet \cite{park2018arbitrary} & 0.034 & - \\  \hline
 Our &0.045  & 0.371 (Region-based)\\
\hline
\end{tabular}
\end {center}
\caption{Execution time comparison (in seconds).}
\label{table:excution_time}
\end{table}
\begin{figure}[htb]
    \centering
    \includegraphics[width=\linewidth]{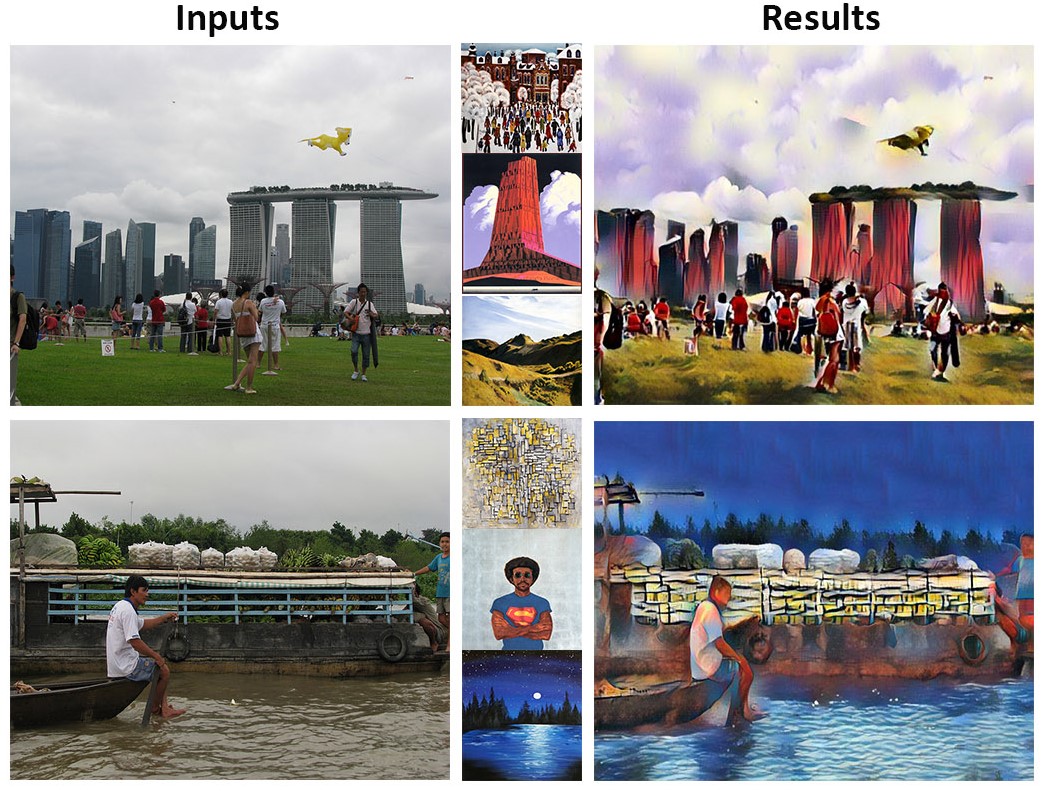}
    \caption{MST with more style references. Our Style Mixer is able to potentially handle arbitrary number of style references.}
    \label{fig:triple}
\end{figure}
\begin{figure}[htb]
    \centering
    \includegraphics[width=\linewidth]{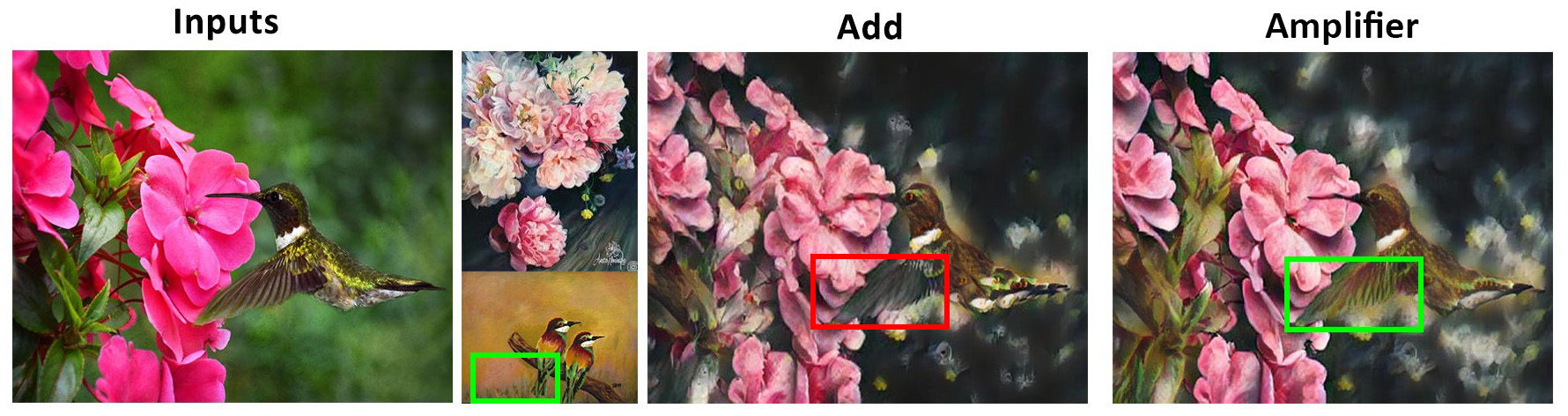}
    \caption{Comparison between add operation and Amplifier as the merging module in calculation of identity loss.}
    \label{fig:amp}
\end{figure}
\subsection{Efficiency}
A run time evaluation has also been conducted, and the results are displayed in Tab.~\ref{table:excution_time}. All the inputs are rescaled to 512 px $\times$ 512 px. In SST, due to the adoption of PA, our model is slightly slower than SANet \cite{park2018arbitrary}, but is still very competitive compared to WCT \cite{li2017universal} and Avatar Net \cite{sheng2018avatar}. In terms of MST, our region-based feature fusion strategy can run at near real-time speed, faster than WCT \cite{li2017universal} but slower than AdaIN \cite{huang2017arbitrary} due to the expensive cost of clustering.
\begin{figure*}[htb]
    \centering
    \includegraphics[width=\linewidth]{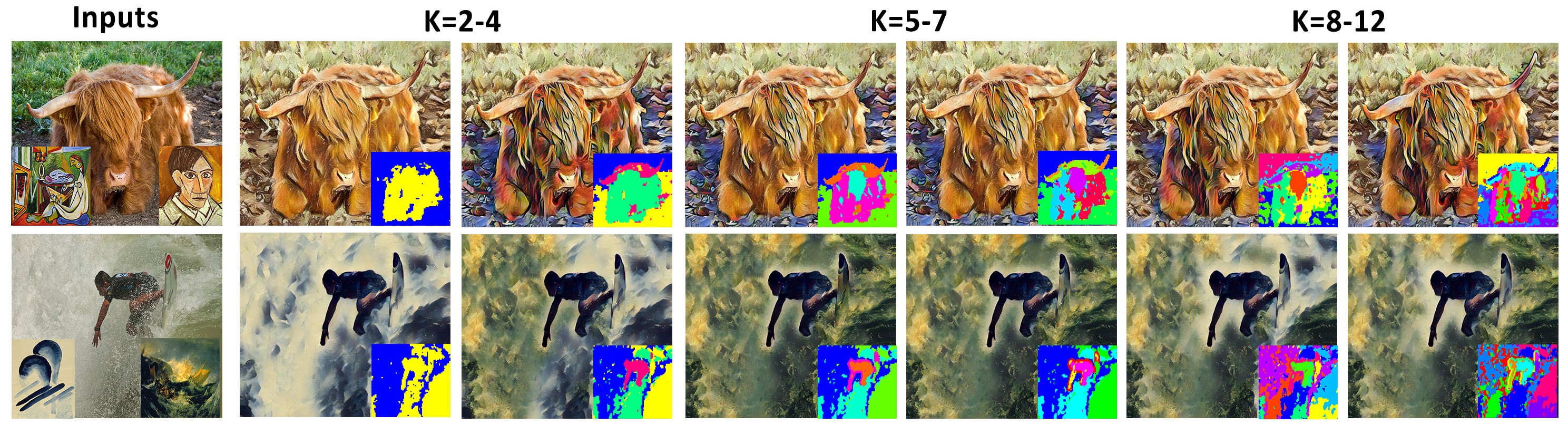}
    \caption{Investigation on different number of clusters.}
    \label{fig:cluster}
\end{figure*}
\subsection{Results with More References} 
Fig.~\ref{fig:triple} shows examples of MST with three references. Our region-based strategy is able to assign different styles to appropriate regions according to semantic correspondence and potentially handle an arbitrary number of references.

\section{Discussion}
\subsection{The Motivation of Amplifier}
\label{sec:dis_amp}
 \cite{park2018arbitrary} introduces identity loss to improve the content preservation and matching ability of style-attention module. When calculating identity loss, SANet merges content feature $F_{c}$ with swapped style feature $F_{s}$ by $F_{cs}=F_{c}+F_{s}$, which is same as normal inference process. However, $F_{c}$ has already contained the necessary information to complete the reconstruction. The chances are that although the network is capable of rebuilding the image, the weights of the attention module is wrongly trained to be 0, which means it makes no effect at all. To solve this vulnerability, we apply Eq. \ref{eq:amp} to replace the original add operation. Without the supply of content image, the PA module is confronted with a bigger challenge and forced to learn more accurate correspondence, which is corroborated by experiments. For example, in Fig.~\ref{fig:amp}, with add operation as merging module, the wings of the bird are wrongly match with the background of flower reference. On the other hand, when the amplifier is being utilized, the wings of the birds are transferred to green color in accordance with that of bird reference.

\subsection{Choice of the Number of Clusters}
To investigate how the number of clusters (K) affects the MST results, we carry out experiments with various content-style pairs, two of which are shown in  Fig.~\ref{fig:cluster}. The experimental results illustrate that the quality of the synthesized result is not sensitive to the size of K when K lies in a restricted range. Typically, K with a size between 5 to 7 inclines to produce appealing results. When K is relatively large, content image is segmented into smaller regions possessing similar characteristics, which are very likely to be assigned with the same style. However, if we further increase the K, unpleasant patterns will occur since small segments are easily influenced by local features and noises, thus producing false matching. It should also be noted that when K is set to a small number, the results are sensitive to the initial seeds of K-means and are not consistent with the semantic information of the content image. 

\subsection{Limitation}
\textbf{Semantic mismatch.} The phenomenon can be attributed to the limitation of the encoder. Since VGG-19 is pretrained on ImageNet, which may not be able to handle the objects that are beyond the predefined categories. Also, there is a distinct domain gap between photos and paintings. As a consequence, some style patterns may be too abstract for VGG19 to extract accurate semantic information. For example, the cloud in the 4th column of Fig.~\ref{fig:control} is wrongly transformed into the pattern of the ground rather than the cloud in that style reference. We believe the development of a more suitable encoder for style images will help to alleviate the problem.

\textbf{Halos near the boundary.}
The segmentation we applied on features is coarser than segmentation of original image due to the shrinking of size. And this deviation will be amplified by the upsampling process and lead to halos. A progressive fusion strategy may be a good direction to resolve this problem.

\section{Conclusion}
In this work, we propose an advanced style transfer network and efficient region-based multi-style transfer strategy. The proposed patch attention module dramatically elevates the ability of semantic style transfer and is applicable to any current attention-based model. Also, we come up with the first region-based strategy for MST, which is proved to be efficient and is capable of improving the consistency of multi-style transfer. Comprehensive experiments demonstrate that our proposed method is favorable compared to other existing methods.

\section*{Acknowledgement}
We thank the anonymous reviewers for helping us to improve this paper. And we acknowledge to the authors of our image and style examples. This work was partly supported by CityU start-up grant 7200607 and Hong Kong ECS grant 21209119.



{\small
\bibliographystyle{ieee}
\bibliography{egbib}
}

\end{document}